\pdfoutput=1
\documentclass[11pt]{article}

\usepackage[]{acl}

\usepackage{times}
\usepackage{latexsym}
\usepackage[T1]{fontenc}
\usepackage[utf8]{inputenc}
\usepackage{microtype}
\usepackage{graphicx}
\usepackage{booktabs}
\usepackage{enumitem}
\usepackage{xurl}
\usepackage{subcaption}
\usepackage{xspace}
\usepackage{todonotes}
\usepackage{tabularray}
\usepackage{listings}
\usepackage{array}
\usepackage{bm}
\usepackage{amsmath}
\usepackage[most]{tcolorbox}
\usepackage{caption}
\usepackage{amssymb}
\usepackage{xcolor}          
\usepackage{color, colortbl} 
\usepackage{minted} 
\definecolor{lightgray}{HTML}{F5F5F5} 
\setminted{
    fontsize=\footnotesize, 
    frame=lines, 
    bgcolor=lightgray, 
    autogobble
}

\newcommand{\grafite}{\textsc{{Grafite}}\xspace}
  
\title{\grafite{}: Generative Regression Analysis Framework \\for Issue Tracking and Evaluation}

\author{
    Ja Young Lee, 
    Mírian Silva, 
    Mohamed Nasr, 
    \\
    \textbf{Shonda Witherspoon, 
    Enzo Bozzani, 
    Veronique Demers, 
    }
    \\
    \textbf{Radha Ratnaparkhi,
    Hui Wu, 
    Sara Rosenthal}
    \\
    \texttt{ja.young.lee@ibm.com}\\
  IBM Research - AI 
  }

\begin{document}

\maketitle

 \begin{figure*}[t]
    \centering
\includegraphics[width=\textwidth]{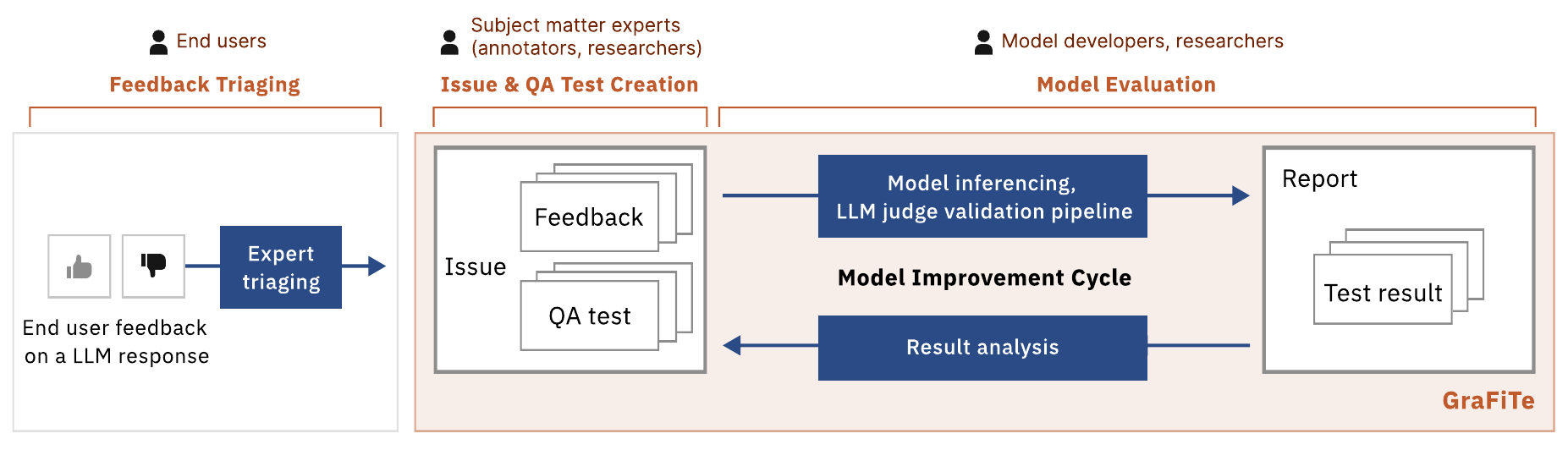}
\caption{The pipeline of \grafite{} and stakeholders of each step}
\label{fig:pipeline}
\end{figure*}

\begin{abstract}
Large language models (LLMs) are largely motivated by their performance on popular topics and benchmarks at the time of their release. However, over time, contamination occurs due to significant exposure of benchmark data during training. This poses a risk of model performance inflation if testing is not carefully executed. To address this challenge, we present \grafite{}, a continuous LLM evaluation platform through a comprehensive system for maintaining and evaluating model issues. Our approach enables building a repository of model problems based on user feedback over time and offers a pipeline for assessing LLMs against these issues through quality assurance (QA) tests using LLM-as-a-judge. The platform enables side-by-side comparison of multiple models, facilitating regression detection across different releases. The platform is available at \url{https://github.com/IBM/grafite}. The demo video is available at \url{www.youtube.com/watch?v=XFZyoleN56k}.
\end{abstract}

\section{Introduction}

Large language models (LLMs) have demonstrated remarkable performance through rapid development as showcased by popular benchmarks. However, there have been growing concerns about the risk of inflated model performance metrics due to benchmark data contamination during training if testing procedures are not meticulously designed and executed~\cite{deng2023investigating, xu2024benchmarkdatacontaminationlarge, zhang2024careful}. There have been various efforts to mitigate this risk by constantly updating the benchmark with newly released datasets \cite{white2024livebench, jain2024livecodebench}, leveraging LLMs~\cite{chatareana_zheng2023} or human intervention~\cite{chiang2024chatbot}. Regardless, challenges remain in tracking a model's regression performance or improvement in a specific domain over time. Constantly updating the benchmark dataset is not well-suited for quantifying absolute performance improvements over time and leveraging LLM or human intervention can introduce subjectivity or bias as the evaluation quality relies on the evaluator \cite{wang-etal-2024-large-language-models-fair}.

Continuous model evaluation in a specific domain presents additional challenges in user experience. Frameworks exist that offer streamlining of LLM evaluation, such as LLMeBench \cite{Dalvi2023LLMeBenchAF}, LM Evaluation Harness \cite{lm_eval_harness}, HELM \cite{heml_liang2023}, and OpenAI Evals\footnote{https://github.com/openai/evals}. While they offer high flexibility with the dataset and evaluators, they focus on one-time performance benchmarking and require technical knowledge to be utilized effectively, as they are command-line interface tools. Tracking model performance using one's own setup remains the responsibility of the user. In addition, while most benchmarks aim to evaluate models' general and broad-use capabilities, they often overlook domain-specific scenarios for specific use cases \cite{anjum2025_domain_specific_bench_eval, McIntosh_2026_inadequacies_bench}. Such scenarios require a nuanced evaluation approach, which standard benchmarks typically do not address.

\begin{table}[t]
    \centering
    \small
    \begin{tabular}{p{7.3cm}}
\toprule
\textbf{Issue Title}: Math - Geometry \\
\midrule
\textbf{User input}: A triangle has a base of 6.5 meters and a height of 4.3 meters. Calculate its area.\\
\midrule
\textbf{Model Output}: \textit{<to be generated by the selected model>}.\\
\midrule
\textbf{Sample Reference Answer (ground truth)}: \\
\midrule
To calculate the area of a triangle, you can use the formula $\text{Area} = \frac{1}{2} \times \text{base} \times \text{height}$. Since the given base is $6.5$ meters and the given height is $4.3$ meters, the area of the triangle is $\frac{1}{2} \times 6.5 \times 4.3$. Thus, $\text{Area} = \frac{1}{2} \times 6.5 \times 4.3 = 3.25 \times 4.3 = 13.975$ square meters. Therefore, the area of the triangle with a base of $6.5$ meters and a height of $4.3$ meters is $13.975$ square meters.\\
\midrule
\textbf{Evaluation Criteria} \\
\midrule
\textbf{Judge Type}: "Prompt template 1: input + output + ground truth + judge guideline"\\
\midrule
\textbf{Judge Guideline:}
1. Use the correct geometric equation to solve.\\
2. In math, the answer must match the reference exactly to be correct.\\
3. The reference and predicted answer should be in the simplest form.\\
4. The input should be a geometry math question.\\
\bottomrule
    \end{tabular}
    \caption{A sample test from an issue.}
    \label{tab:example}
    \vspace{-1.4em}
\end{table}

To address these challenges, we present a web-based platform, \grafite{}, for continuous LLM benchmarking through issue creation and management with model evaluation using an ensemble of LLM judges~\cite{verga2024replacing} and human-in-the-loop annotation. The \grafite{} pipeline is shown in Figure~\ref{fig:pipeline}: When an end user asks a question and is unsatisfied, they provide a thumbs down as feedback. This triggers expert triaging by subject matter experts who use \grafite{} to assess whether an issue and test should be created. All issues can then be evaluated by LLM judge validation to generate a report as feedback to model developers. This analysis can also be used to improve and adjust the issues and tests as needed. An example of an issue and test is shown in Table~\ref{tab:example}. This design enables end users to contribute their feedback and provides model developers and researchers with detailed performance reports while maintaining the integrity of the data. Our contributions are: 
\begin{itemize}[leftmargin=*]
    \item \grafite{} issue tracking: A novel framework to track domain-specific model problems across model versions.  Our platform serves as a repository of model issues, associated with human-contributed tests per issue, to validate model performance on each issue.
    \item \grafite{} evaluation:  By leveraging human-contributed test sets, researchers and model developers can assess model performance over time across versions or against other models, gaining insights into model improvements and regressions within targeted scenarios.
By evaluating models on human-created test sets that target specific use cases not visible to model developers, \grafite{} supports more robust model development by mitigating the risk of inflated performance that can result from focusing on popular benchmarks.
    \item \grafite{} analytics: An analytics visualization tool that facilitates understanding of model performance. This feature streamlines benchmarking models across domains, versions, and comparisons between different models through a user-friendly graphical user interface (GUI).
    \item We accompany \grafite{} with a use case study on sample data with a publicly available model to fully illustrate its capabilities and a user study to assess the platform's efficacy and usability. We open-source our platform  to the community, along with the sample data, at \url{https://github.com/IBM/grafite} 
    to promote collaborative research advancement. 
\end{itemize}


\section{\grafite{} Platform}
\grafite{} is a web application developed using Next.js\footnote{https://nextjs.org/}, Ag-grid\footnote{https://www.ag-grid.com/}, Carbon Design System\footnote{https://carbondesignsystem.com/}, RabbitMQ\footnote{https://www.rabbitmq.com/}, FastAPI\footnote{https://fastapi.tiangolo.com/}, and MongoDB\footnote{https://www.mongodb.com/}. It is compatible with Ollama\footnote{https://ollama.com/} and WatsonX.ai\footnote{https://www.ibm.com/products/watsonx-ai} for model inference. Users can easily execute tasks using the GUI. All data, including evaluation results, is stored in MongoDB for issue management and long-term performance tracking.

The system is designed to enable model developers and researchers to track and manage selected model problems and evaluate LLMs over time. As shown in the shaded box in Figure \ref{fig:pipeline}, the app aggregates model problems into a set of issues and tests. These issues are assessed using the LLM-as-a-judge technique \cite{chatareana_zheng2023}, which translates qualitative evaluations into quantitative metrics using LLMs as a panel of evaluators \cite{verga2024replacing}. This LLM-based evaluation is combined with human evaluation to streamline the analysis of model performance and generate reports.

\subsection{Issues and Tests}
\label{sec:issue_tests}

The \textbf{issue} is the bedrock of the \grafite{} system. An issue is the erroneous behavior of a model during testing. It is synthesized from multiple sources of domain-specific feedback. Each issue includes metadata concerning: (a) a detailed description of the identified error, (b) user feedback from which the issue originated for reference, (c) resolution statuses, (d) domain tags, and (e) a set of tests. 

A \textbf{test} is an example that exhibits the erroneous behavior of the issue. It is composed of an input prompt designed to be submitted to the model; a reference answer when applicable; and an LLM-as-a-judge evaluation configuration (e.g., whether to do reference-guided grading or single answer grading~\cite{chatareana_zheng2023}). The judge component relies on a structured prompt template and scoring guidelines to ensure consistency and task-aware assessment. See Appendix \ref{appendix:judge_templates} for judges' prompt templates reference. Tests and issues are manually created to challenge models beyond benchmark routines.


\subsection{Evaluation}
\label{sec:evaluation}
\grafite{} employs an LLM-as-a-judge approach to enable quantitative analysis of qualitative assessments. Each test is scored on a binary scale of 0 or 1, with accompanying justifications or rationales. The judge uses a structured prompt template that guides the LLM to serve as an impartial judge and to assign a score based on selected inference data and the generated response. It also includes instructions for scoring criteria, a standardized output format, and user-provided rating guidelines as described in Appendix~\ref{appendix:judge_templates}. 
To mitigate the potential bias or quality issues associated with the LLM-as-a-judge technique \cite{wang-etal-2024-large-language-models-fair,verga2024replacing}, \grafite{} incorporates an ensemble of LLM-judges and determines whether a test passes or fails based on the average. Users have the flexibility to configure the judge models according to their requirements. 


 \begin{figure*}[t]
    \centering
\includegraphics[width=\textwidth]{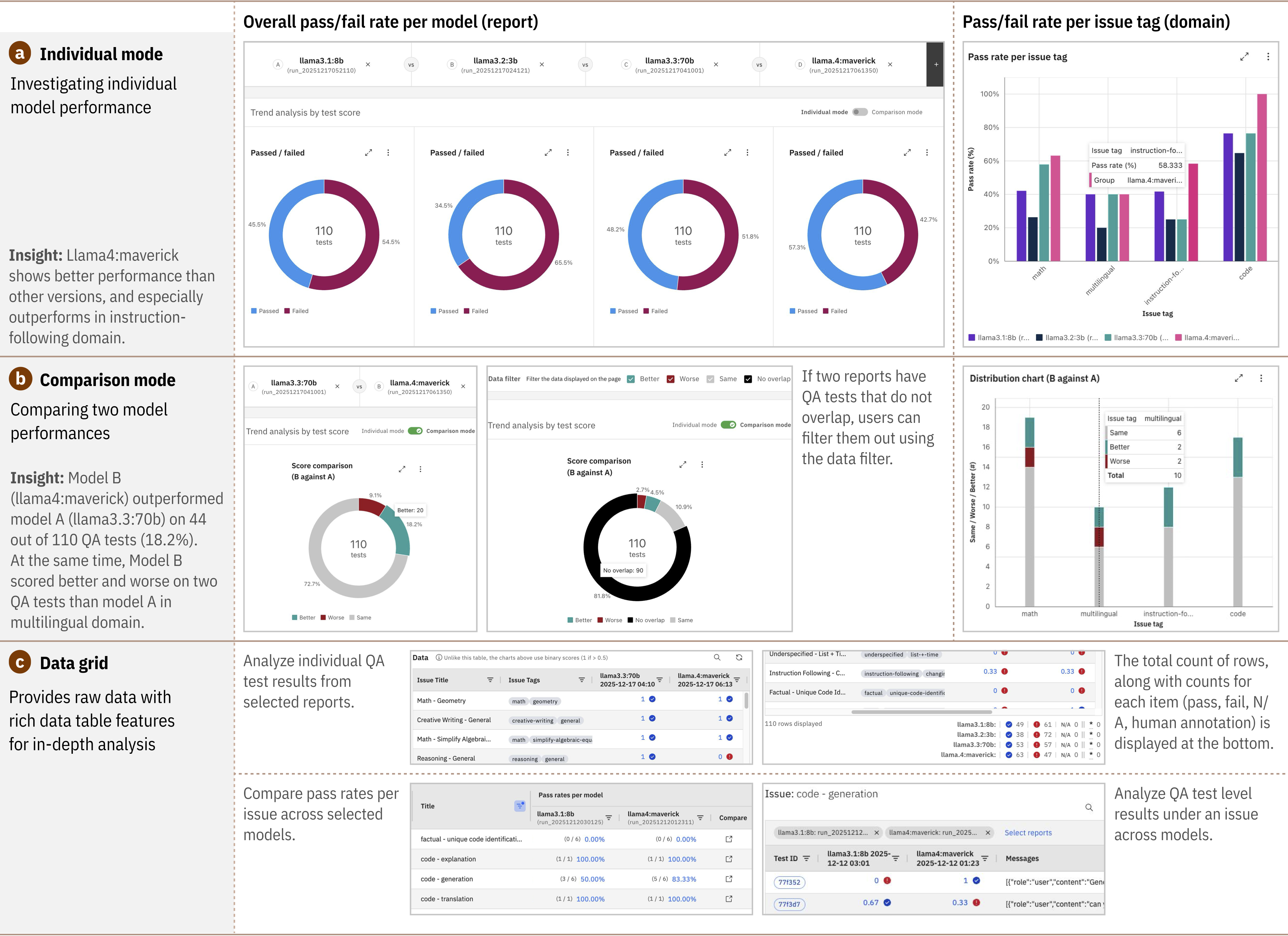}
\caption{Illustration of \grafite{}'s evaluation components. Screenshots are drawn from the use case test run results, described in Section \ref{sec:use case} }
\label{fig:evaluation}
\end{figure*}

The model evaluation for \grafite{} is conducted by analyzing scores from LLM judges, complemented by human input where necessary. When the judge model's output is inadequate, users can add a human evaluation, overriding the judge's score and justification. By monitoring quantified binary scores per issue, users gain a succinct overview of model deficiencies and any potential performance regressions across different versions over time. This is explained in more detail with illustrative examples in Section \ref{sec:use case}.

\subsection{Report Generation}

\grafite{} offers two interfaces for analyzing the output of model evaluations: (1) test report and (2) trend analysis. The test report supports the evaluation of an individual model by analyzing the overall test run results and individual test scores. Trend analysis enables comprehensive assessments of trends for a single model or across multiple models. Comparing multiple models is a useful feature for exploring how model performance has changed over time across different versions of the same model, for comparing the performance of model families, or comparing competitive models.

\section{\grafite{} Workflow}

Users go through four primary steps for model evaluation in \grafite{}: (1) issue management (lookup and creation), (2) test creation, (3) test run execution, and (4) model evaluation. Issues and tests are reusable for multiple assessment cycles. 

\subsection{Issue Management}
The workflow begins with a manual \textit{issue} validation step to ensure uniqueness, enabling reuse of an existing issue or creation of a new entry. Each issue encapsulates a specific erroneous behavior to be monitored, so this manual search is based on behavior similarity. Users can use issue tags for efficient filtering of relevant issues. When a matching issue is identified, new tests could be associated with the existing entry. Otherwise, a new issue is created and the user is required to summarize the behavior and associate the supporting feedback that evidences the identified error. 


\subsection{Test Creation}
Once the issue is ready, users proceed to create a set of tests designed to validate the identified problem. These tests should cover diverse facets of the issue to ensure thorough evaluation. For example, to test deficiencies in mathematical calculations, users can create multiple tests that vary in mathematical complexity (See example in Table~\ref{tab:example}). Additional tests can also be added later, particularly if new feedback from model responses is received. 

The test creation process also includes an evaluation setup, where the user selects the most appropriate judge template and provides judge scoring criteria. These can also be inherited from other tests in the issue and then tailored specifically to the current test. See Appendix \ref{appendix:judge_guideline} for judges' evaluation guideline examples.

\subsection{Test Run Execution}
\grafite{} automates model inference and judgment evaluation of multiple tests through its pipeline. Users select a subset of issues and/or tests from the list to be executed by the model. For example, the user can filter tests using the tags to only get issues of a particular domain (e.g., math). 
\grafite{}’s test run pipeline first generates model outputs for the chosen tests. Subsequently, the pipeline conducts LLM-as-a-judge evaluations using each test’s judge configuration. Users have the option to conduct a single-model judge evaluation or an ensemble of judges. Each test run generates a report with details of the evaluation.  

\subsection{Model Analysis}

The user explores the test report. They can see a pie chart summarizing pass or fail rates and detailed test outcomes. In the case of an ensemble of judges, a test passes if the average score exceeds 0.5. The main section of the page provides an in-depth examination of individual unit test outcomes, presenting judge scores alongside justifications. Users can assess results individually or at the issue level to gauge model performance within specific domains. Each issue’s failure rate alerts users to potential areas that require attention. When the judge's result is insufficient, users can add a human evaluation. It overrides the judge score, as described in the previous section, to enhance the result more robustly by not relying on the quality of judge LLMs. 

The trend analysis page offers the user a visualization dashboard that facilitates multi-model performance benchmarking, as shown in Figure \ref{fig:evaluation}. Upon selecting reports, users see two charts and a data grid listing test results. Users can compare model performance based on pass rates and across different issue tags or domains. The dashboard offers two modes: (a) Comparison mode and (b) individual mode (Figure \ref{fig:evaluation}a, \ref{fig:evaluation}b). In comparison mode, two reports are selected and each chart illustrates how many tests from one model outperform, underperform, or match another’s score (Figure \ref{fig:evaluation}b). The data grid with rich functionalities streamlines in-depth data analysis (Figure \ref{fig:evaluation}c). This dashboard enables users to rapidly assess performance disparities between models and across domains.

\section{Usage}
\label{sec:use case}

\begin{table}[!t]
\rowcolors{2}{gray!15}{white}
\footnotesize
\centering
\setlength{\tabcolsep}{3pt}
\begin{tabular}{p{1.8cm} p{4.3cm}cc}
\toprule
\textbf{Domain} & \textbf{Issues (Task Types)} & \textbf{N$_i$} & \textbf{N$_t$} \\ 
\midrule
Code & Fixing/Modifying, Generation, Optimization & 3 & 17 \\
Creative & General Writing & 1 & 6 \\
Factual & Explanation/Description, Factoid Entity, Unique Code Identification & 3 & 17 \\
Instruction Following & Changing Input, Extract Information & 2 & 12 \\
Math & Geometry, Riddle With Manipulation, Simplify Algebraic Equations, Word Pattern & 4 & 20 \\
Multilingual & No Translation Needed, Translation & 2 & 10 \\
Reasoning & General & 1 & 6 \\
Summarization & Summarization + Explanation & 1 & 6 \\
Table & Calculation, Modification & 2 & 11 \\
Underspecified & List + Time & 1 & 5 \\ 
\midrule
\textbf{Total (10)} & & \textbf{20} & \textbf{110} \\
\bottomrule
\end{tabular}
\caption{Issue distribution aggregated by domain. $N_i$: number of issues. $N_t$: number of tests.}
\label{tab:data_stats}
\vspace{-1.4em}
\end{table}

\grafite{} supports two core functionalities: (1) tracking and managing domain-specific model issues and (2) evaluating models and analyzing performance changes. To illustrate the applicability of these functionalities, we conducted an assessment using a sample dataset on open-source models.

In Table \ref{tab:data_stats}, we aggregate the tasks that compose the issues across our 10 domains, showing how we categorize error behaviour across tests. In total, the dataset used in this assessment comprises 20 issues and 110 tests, combining human-generated instances and curated benchmark data from Chatbot Arena \cite{chatareana_zheng2023}. This distribution ensures coverage of both structured reasoning tasks (e.g., Math and Code) and open-ended generation tasks (e.g., Creative and Summarization). One representative test per issue (task type) is publicly available at \url{https://github.com/IBM/grafite/blob/main/server/seed}. 


We evaluate open-source Meta Llama\footnote{https://huggingface.co/meta-llama} models across four versions: \textit{Llama-3.1-8B-Instruct}, \textit{Llama-3.2-3B-Instruct}, \textit{Llama-3.3-70B-Instruct}, and \textit{Llama-4-Maverick-17B-128E-Instruct} \cite{grattafiori2024llama3herdmodels, abdullah2025evolutionmetasllamamodels}. We used three judge models as an ensemble of judges: \textit{Llama-3.3-70B-Instruct}\footnote{There is a known preference leakage issue \cite{li2025preference} with using the same model or the same model family for both inference and evaluation using the LLM-as-a-judge technique; however, the primary objective of this evaluation is to demonstrate the capability and utility of \grafite{}, so such a caveat may be reasonably disregarded in this context.} \cite{grattafiori2024llama3herdmodels}), \textit{Phi-4} from Microsoft\footnote{https://huggingface.co/microsoft/phi-4} \cite{abdin2024phi4technicalreport}, \textit{Gpt-oss-120b} from OpenAI\footnote{https://huggingface.co/openai/gpt-oss-20b} \cite{openai2025gptoss120bgptoss20bmodel}. The results are described in Table \ref{tab:use_case}, where we present the models' pass rates aggregated by issue domain. Our assessment demonstrates how users can perform continuous evaluations of model iterations through \grafite{}.


\begin{table}[t]
\rowcolors{2}{gray!15}{white}
\footnotesize

\setlength{\tabcolsep}{3pt} 
\begin{tabular}{>
{\raggedright\arraybackslash}p{2.5cm}>
{\raggedright\arraybackslash}p{0.9cm}>
{\raggedright\arraybackslash}p{0.9cm}>
{\raggedright\arraybackslash}p{1.1cm}>
{\raggedright\arraybackslash}p{1.3cm}}

\toprule
\rowcolor{white} \textbf{Issue Domain}
& \textbf{L3.1} 
& \textbf{L3.2} 
& \textbf{L3.3} 
& \textbf{LMav} \\ 
\rowcolor{white} \textbf{($n$ tests)}
& \textbf{8B} 
& \textbf{3B} 
& \textbf{70B} 
& \textbf{17B} \\ 

\midrule

Total (110)           & 42.1  & 26.3  & 57.9  & 63.2  \\
\midrule
Code (17)             & 76.5  & 64.7  & 76.5  & 100   \\
Creative (6)          & 100   & 83.3  & 100   & 83.3  \\
Factual (17)          & 5.9   & 5.9   & 17.7  & 29.4  \\
Instr. follow (12)    & 41.7  & 25    & 25    & 58.3  \\
Math (20)             & 40.0  & 25.0  & 55.0  & 60.0    \\
Multilingual (10)     & 40.0  & 20.0  & 40.0  & 40.0    \\
Reasoning (6)         & 66.7  & 0     & 83.3  & 33.3  \\
Summarization (6)     & 83.3  & 83.3  & 66.7  & 83.3  \\
Table (11)            & 36.4  & 44.5  & 36.4  & 54.5  \\
Underspec. (5)        & 0.0    & 16.7  & 0.0  & 0.0     \\

\bottomrule
\rowcolor{white}\multicolumn{5}{p{0.95\columnwidth}}{\scriptsize \textit{*All column values are mean values from an ensemble of judges' scores.}}
\end{tabular}
\caption{Llama Models Pass rate (\%) per issue.}
\label{tab:use_case}
\end{table}

The following are the primary insights from this evaluation\footnote{Detailed results of this evaluation with the sample data: https://github.com/IBM/grafite/}
First, a regression is observed between \textit{Llama-3.1} and \textit{Llama-3.2}, which may be attributed to 3.2's smaller parameter size. Notably, the \textit{Llama-3.2} model underperformed particularly on reasoning tests, indicating that it lacks factual knowledge. This indicates that the model is unsuitable for such purposes or requires alternative approaches, such as fine-tuning or tool-calling. Second, all versions performed poorly in an "underspecified" domain, where models are expected to request additional context before answering. However, their failure reasons differ. \textit{Llama-4} tends to generate code to retrieve information without being instructed to do so. \textit{Llama-3.1} and \textit{Llama 3.2} generate comprehensive lists that often include hallucinations. Meanwhile, \textit{Llama-3.3} suggests ways for users to obtain the information themselves, while noting it cannot provide real-time data. Depending on the use case, some responses may be acceptable, which could be reflected by updating the test judge guidelines for this domain.


\section{User Study}

To validate the efficacy of \grafite{}, we conducted a user study to assess usability and how well the app's features support model performance evaluation. To identify which aspects most contribute to user satisfaction, we surveyed participants about their experiences across multiple dimensions using a 5-point Likert scale.


\textbf{Demographic information:} The participants pool (eight in total) included software engineers, product managers, AI/ML researchers, and data analysts. In terms of experience, participants demonstrated varying levels of familiarity, ranging from “moderately familiar” to “highly experienced” with AI/LLM expertise and AI model evaluation. \grafite{} usage tenure ranged from less than one month to more than six months, reflecting a mix of both new and established users.

\begin{table}[!t]
    \small
    \rowcolors{2}{gray!15}{white}
    \begin{tabular}{p{5.5cm} c c }
        \toprule
        \textbf{Platform Feature/User Experience Dimension} & $\bm{\mu}$ & $\bm{\sigma}$ \\
        \midrule
        Value of human-in-the-loop annotation for improving evaluation quality & 4.13 & 0.64 \\
        Usefulness of ensemble evaluation approach (multiple LLM judges) & 4.13 & 0.84 \\
        Usefulness of pass/fail metrics for evaluating model performance & 4.13 & 0.84 \\
        Effectiveness in tracking regression tests across issues & 4.13 & 0.84 \\
        Confidence in LLM-as-a-judge evaluation accuracy & 4.00 & 0.53 \\
        Ease of creating test runs & 4.00 & 0.76 \\
        Satisfaction with speed of performing common tasks & 3.88 & 0.64 \\
        Overall satisfaction with GRAFITE & 3.88 & 0.84 \\
        Clarity and understandability of generated reports & 3.75 & 0.71 \\
        Ease of creating new issues & 3.63 & 0.74 \\
        Ease of linking tests to issues & 3.25 & 0.71 \\
        Ease of navigating the interface & 3.00 & 0.76 \\
        Ease of organizing and categorizing issues & 3.00 & 0.76 \\
        \bottomrule
    \end{tabular}
    \caption{User Experience Statistics for \grafite{}: Dimensions ranked by average satisfaction and effectiveness ratings. Higher $\bm\mu$ values indicate greater user satisfaction, while lower $\bm\sigma$ values indicate more consistent agreement among participants.}
    \label{tab:ux_stats}
    \vspace{-1.4em}
\end{table}

                                            
Table \ref{tab:ux_stats} presents the average ratings that summarize the key platform features. \grafite{} demonstrates strong core performance and product-market fit, with 87.5\% of participants recommending the platform to colleagues. Results revealed strong appreciation for \grafite{}'s core evaluation capabilities, with participants highly valuing the human-in-the-loop annotation feature ($\mu$ = 4.13), the ensemble evaluation approach with multiple LLM judges ($\mu$ = 4.13), pass/fail metrics ($\mu$ = 4.13), visual analytics and charts ($\mu$ = 4.33), and the effectiveness of regression test tracking ($\mu$ = 4.13). Participants also expressed confidence in the platform's LLM-as-a-judge evaluation accuracy ($\mu$ = 4.00), found creating test runs straightforward ($\mu$= 4.00), and reported solid overall satisfaction ($\mu$ = 3.88). 

Alongside the highly rated aspects, certain dimensions showed moderated mean scores, suggesting opportunities for refinement, such as interface navigation ($\mu$ = 3.00) and issue organization ($\mu$ = 3.00). There is room to improve the platform's GUI for broader adoption including automatic detection of underperforming issues, visual heatmaps of performance metrics, and a customizable dashboard with baseline model data.






\section{Conclusion}

In this paper, we propose a novel framework, \grafite{}, designed to track and evaluate language model performance problems, grouped into domain-specific issues. Our platform enables automated test runs and human-in-the-loop verification for model benchmarking across domains and providers. Our experiments with open-source models illustrate our design's practical application. The platform features, especially evaluation capabilities, are beneficial and play a key role in assessing model performance from release to release. In our user study, we showed that our app is ideal for enterprise use. In our future work, we plan to develop more adaptable issue categorization tools, to enhance the efficacy of issue tracking in large-scale issue management. We will also explore automating repetitive model evaluation tasks through sophisticated approaches, such as augmenting test sets through synthetic data generation.

\section{Ethical Considerations}

The \grafite{} platform is intended for monitoring LLMs on human-generated issues and tests. The platform itself does not have any limitations on what kind of data can be stored and represented in the platform. Evaluation in \grafite{} is primarily through LLM-as-a-Judge and therefore subject to biases of the LLMs used. While we mitigate this by taking an ensemble approach, there is still likely to be biases and incorrect evaluations. Users should analyze evaluation results with the understanding that it should be used to observe trends and spot areas to explore further, even though biases and incorrect evaluations may exist. This understanding should also be used to analyze the results reported in the paper.

\bibliography{references}
\clearpage

\onecolumn

\appendix

\section{LLM-as-a-Judge Prompt Templates}
\label{appendix:judge_templates}

As introduced in Section \ref{sec:evaluation}, in our evaluation approach \grafite{} employs an LLM-as-a-judge setup enabling quantitative analysis of qualitative assessments. The evaluation configuration for our LLM judges is composed of a judge prompt template and judge evaluation guidelines. The judge prompt template can be selected from three options, where each indicates which components from the inference data should be evaluated: (a) input, model output, and ground truth; (b) model output and ground truth, or (c) input and model output. This optimization is crucial to minimize resource consumption during test run pipeline execution. For example, when evaluating a task to generate fewer than 100 words, only the model’s output may be necessary, as there are many appropriate responses. Additionally, the evaluation criteria must balance precision and nuance, ensuring that minor variations are neither penalized nor accepted as correct. In Figure \ref{fig:judge_templates}, we describe the full prompt of the three judge templates used in our settings.  

\begin{figure}[!ht]
    \fbox{
        \begin{minipage}[t][8.5cm][t]{0.485\textwidth}
        \small
        Act as an impartial judge and evaluate the text generated by an AI model displayed below. Assign a score using a binary 0/1 scale based on the guidelines provided.\\
        
        **Input to the model**\\\{\{prompt\_text\}\}
        
        **Model output to be rated**\\
        \{\{model\_response\}\}
        
        **Ground truth text**\\
        \{\{ground\_truth\}\}
        
        **Rating Guidelines**\\
        The model output should adhere to the following guidelines:\\
        \{\{judge\_guidelines\}\}
        
        **Scoring Criteria**\\
        * Score 0: The model output violates any of the guidelines.
        * Score 1: The model output is well aligned with the ground truth, the input to the model, and adheres to all guidelines.
        
        **Return Your Rating**\\
        Return your rating in the following format:
        \{"justification": your\_justification, "score": your\_score\}\\
        **Your rating:**
        \subcaption{\textbf{Judge Prompt template 1: input + output + ground truth + judge guideline.}}
        \end{minipage}
    }
    \hfill
    \fbox{
        \begin{minipage}[t][8.5cm][t]{0.483\textwidth}
        \small
        Act as an impartial judge and evaluate the text generated by an AI model displayed below. Assign a score using a binary 0/1 scale based on the guidelines provided.\\

        **Model output to be rated**\\\{\{model\_response\}\}
        
        **Ground truth text**\\\{\{ground\_truth\}\}
        
        **Rating Guidelines**\\The model output should adhere to the following guidelines:
        \{\{judge\_guidelines\}\}
        
        **Scoring Criteria**\\        
        * Score 0: The model output violates any of the guidelines.
        * Score 1: The model output is well aligned with the ground truth, the input to the model, and adheres to all guidelines.
        
        **Return Your Rating**
        
        Return your rating in the following format:
        \{"justification": your\_justification, "score": your\_score\}
        
        **Your rating:**\\
        \\
        \subcaption{\textbf{Judge Prompt template 2: output + ground truth + judge guideline.}}
        \end{minipage}
    }
    \hfill
    \fbox{
        \begin{minipage}{1\textwidth}
        \small
        Act as an impartial judge and evaluate the text generated by an AI model displayed below. Assign a score using a binary 0/1 scale based on the guidelines provided.
        
        **Input to the model**
        
        \{\{prompt\_text\}\}
        
        **Model output to be rated** 
        
        \{\{model\_response\}\}
        
        **Rating Guidelines**
        
        The model output should adhere to the following guidelines:
        \{\{judge\_guidelines\}\}
        
        **Scoring Criteria**
        * Score 0: The model output violates any of the guidelines.
        * Score 1: The model output is well aligned with the ground truth, the input to the model, and adheres to all guidelines.
        
        **Return Your Rating**
        
        Return your rating in the following format:
        \{"justification": your\_justification, "score": your\_score\}
        
        **Your rating:**"
        \subcaption{\textbf{Judge Prompt template 3: input + output + judge guideline.}}
        \end{minipage}
    }
    \caption{Judge Prompt Templates: Options available for the user in the unit test evaluation settings.}
    \label{fig:judge_templates}
\end{figure}

\subsection{Judge Evaluation Guidelines per Domain Examples}
\label{appendix:judge_guideline}

For each test, the user should provide the judge prompt template, as already described in Appendix \ref{appendix:judge_templates}, and within each prompt, the model output should adhere to the provided guidelines. These guidelines are domain-specific and should correlate with the type of task. To illustrate, Figure ~\ref{fig:creative_guideline_example} has guideline examples for the Code, Creative, and Factual issue domains. These guidelines are examples based on the issue domains for our use-case study (Section \ref{tab:use_case}); the researcher and developer or test creator can define the guidelines as desired. 

\begin{figure}[!t]
    \fbox{
        \begin{minipage}[t][3.5cm][t]{0.485\textwidth}
        \small
        1. In creative writing, there is no reference answer.\\
        2. Correct answers can be of different topics and lengths, unless the question specifies otherwise.\\
        3. If instructions are provided in the question, they must be followed.\\
        4. Answers that are incomplete or have misspellings should be penalized.\\
        5. Input should be a request for creative writing.\\
        \subcaption{\textbf{Judge guideline for \textit{Creative} domain issue.}}
        \end{minipage}
    }
    \hfill
    \fbox{
        \begin{minipage}[t][3.5cm][t]{0.485\textwidth}
        \small
        1. Entity in question should be connected to a specific topic as described in input.\\
        2. Response should contain only factual information.\\
        3. Response should confirm information in question when it isn't clear\\
        4. The question should have details pertaining to that specific entity.\\
        \\
        \subcaption{\textbf{Judge guideline for \textit{Factual} domain issue.}}
        \end{minipage}
    }
    \fbox{
    \begin{minipage}[t]{1\textwidth}
        \small
        1. Model output should be consistent and well aligned with the desired output.\\
        2. The generated code must strictly follow the instructions in the prompt and use the specified programming language.\\
        3. Alternative correct implementations are acceptable if they meet all requirements from the input prompt.\\
        4. The presence of additional explanation or context is allowed and does not make the output wrong or mark it down, as long as it remains relevant to the question and does not contradict the code.\\
        5. The input should be a request to generate code.
        \subcaption{\textbf{Judge guideline for \textit{Code} domain issue.}}
    \end{minipage}
    }
    \caption{Judge Guideline: Evaluation guidelines provided by the user to be part of the LLM-as-a-judge prompt template. Examples of three different issue domains.}
    \label{fig:creative_guideline_example}
\end{figure}

\end{document}